\pdfoutput=1
\documentclass[twocolumn]{ceurart}

\usepackage{subcaption}
\usepackage{threeparttable}
\usepackage{alltt}
\usepackage{tcolorbox}
\tcbuselibrary{skins}

\usepackage{listings}
\newcommand{\hlmask}{\texttt{<MASK>}}
\newcommand{\jkey}[1]{\textcolor{blue!75!black}{"#1"}}
\newcommand{\jval}[1]{\textcolor{teal!55!black}{"#1"}}
\newcommand{\jnote}[1]{\textcolor{gray!70}{\normalfont\scriptsize\textit{\ // #1}}}

\newif\ifanonymous
\anonymousfalse

\begin{document}

\copyrightyear{2026}
\copyrightclause{Copyright for this paper by its authors.
  Use permitted under Creative Commons License Attribution 4.0
  International (CC BY 4.0).}

\conference{RecSys in HR 2026: The 6th Workshop on Recommender Systems for
  Human Resources, in conjunction with the 20th ACM Conference on Recommender
  Systems, September 2026, Prague, Czech Republic}

\title{JobHop~v2: A Large-Scale Career Trajectory Dataset from
  Unstructured Resumes}

\ifanonymous
\else
\author{Iman Johary}[%
  orcid=0009-0007-9223-9995,
  email=iman.johary@ugent.be,
]
\cormark[1]
\address{AIDA-IDLab, Department of Electronics and Information Systems,
  Ghent University, Ghent, Belgium}

\author{Guillaume Bied}[%
  email=guillaume.bied@ugent.be,
]

\author{Alexandru C. Mara}[%
  email=alexandru.mara@ugent.be,
]

\author{Tijl De Bie}[%
  email=tijl.debie@ugent.be,
]

\cortext[1]{Corresponding author.}
\fi

\begin{abstract}
  Large-scale, richly annotated career trajectory data underpins workforce
  planning, job recommendation, and labour market analysis, yet publicly
  available datasets are either small, closed to independent use, or built from
  pre-standardized occupational codes with LLM-synthesized rather than authentic
  free text.
  We present \textbf{JobHop~v2}, an improved version of the publicly available
  JobHop dataset, constructed through end-to-end large language model (LLM)
  extraction from a corpus of ${\sim}440{,}000$ pseudonymized, multilingual
  resumes provided by VDAB, the Flemish Public Employment Service.
  The released dataset comprises $355{,}315$ career trajectories annotated with
  ESCO occupational codes, quarter-level temporal information, and normalized
  five-level education attainment, broadening both the coverage and the
  annotation richness of the original release.
  Relative to v1, JobHop~v2 introduces a redesigned extraction pipeline based on
  reasoning-controlled LLM inference with a retry mechanism (achieving a
  $100\%$ JSON parse rate), a richer extraction schema, and a revised evaluation protocol scored
  against three complementary annotation baselines. Evaluated against these
  baselines, our best extractor comes closest to the inter-annotator agreement
  ceiling among all compared models, trailing it by only $1.1$--$2.7$~percentage
  points.
  The dataset and code are publicly released to support reproducible
  career-trajectory research.
\end{abstract}


\maketitle

\section{Introduction}\label{sec:introduction}

Understanding how careers evolve (which roles people transition into, when, and
from what educational background) has broad practical value: it supports
evidence-based labour market policy, powers job recommendation systems, and
enables career counseling at scale~\cite{decorte2023career,karrierewege}.
The limiting factor for computational career analysis is data. Resumes are the
richest naturally occurring source of career-trajectory information: unlike job
postings, which capture only open roles, or administrative records, which rarely
include job details, resumes document the full arc of a working life (job
titles, responsibilities, education, and the timing of transitions) in a single
document. Yet their unstructured, multilingual, and heterogeneous nature has
long prevented large-scale systematic use.

Recent large language models (LLMs) have changed this calculus. By combining
open-ended text comprehension with structured reasoning, LLMs can normalize job
titles to occupational taxonomies, resolve temporal ambiguities, and extract
skill inventories from free-form descriptions at a scale and accuracy that
rule-based and sequence-labeling pipelines could not
approach~\cite{du2024labor,jobhop-v1}. It is now possible to source rich career
trajectory data containing temporal and educational signals from unstructured
resumes, enabling large resume corpora to be used for career-path analysis and
modeling.

Despite this opportunity, the field still lacks a suitable public benchmark.
Existing public datasets are either small (Decorte et
al.~\cite{decorte2023career} release $2{,}164$ career histories and
OpenResume~\cite{openresume} only $301$ real resumes), built from
pre-standardized occupational codes rather than raw text~\cite{karrierewege}, or
derived from proprietary platforms~\cite{vafa2022career,zha2024unitrep,li2017nemo}
that are not publicly available for independent use. JobHop~\cite{jobhop-v1} was, to our
knowledge, the first large-scale public dataset to pair ESCO occupation codes
with both temporal information and educational attainment, extracted end-to-end
from real unstructured resumes. However, its extraction pipeline suffered from
two practical limitations: degraded extraction quality on noisy multilingual
inputs, and failure to produce consistently machine-parseable output.

We present \textbf{JobHop~v2}, an improved version of the publicly available
JobHop dataset~\cite{jobhop-v1}. Starting from ${\sim}440{,}000$ pseudonymized,
multilingual resumes provided by VDAB, reasoning-controlled LLM extraction yields
$355{,}315$ career trajectories annotated with ESCO taxonomy
codes~\cite{le2014esco}, quarter-level temporal information, and five education
levels. The revised pipeline improves on the original release by integrating a
redesigned extraction schema, a reasoning-effort-controlled inference
configuration with a multi-step retry mechanism, a two-step ESCO
occupation-code assignment policy, a five-stage cleaning workflow, and a revised
extraction-evaluation protocol scored against three complementary annotation
baselines.

\noindent This paper makes the following contributions:
\begin{itemize}
  \item \textbf{JobHop~v2}: a public dataset of $355{,}315$ career
    trajectories with ESCO occupation codes, quarter-level temporal
    annotations, and five-level education attainment, an improved release of
    the original JobHop benchmark~\cite{jobhop-v1} with broader annotation scope
    and higher extraction quality.
  \item \textbf{An extraction pipeline} based on reasoning-controlled LLM
    inference with a multi-step retry mechanism, achieving a $100\%$ JSON parse
    rate after retry over ${\sim}400{,}000$ input resumes and, across three
    annotation baselines, coming closest of all compared models to the
    inter-annotator agreement ceiling.
  \item \textbf{A revised evaluation protocol} with graduated partial-credit
    scoring and three complementary annotation baselines, disentangling genuine
    extraction errors from annotation noise.
\end{itemize}

\noindent\textit{Ethics.}
All processing was performed on local infrastructure; resume contents were never
transmitted to external services. The released dataset incorporates
multi-layered privacy protections: VDAB pseudonymization, removal of location
fields, and coarsening of all dates to quarter-level granularity.

\section{Related Work}\label{sec:related_work}

\paragraph{Career trajectory datasets.}
Computational career analysis depends on the quality and scale of available
trajectory data. Proprietary sources such as
LinkedIn~\cite{li2017nemo,meng2019hierarchical}, Randstad~\cite{schellingerhout2022},
Zippia~\cite{vafa2022career,du2024labor}, and large in-house corpora such as
UniTRep~\cite{zha2024unitrep} ($1.19$M trajectories) and
TAPJFNN~\cite{qin2020tapjfnn} provide extensive longitudinal histories but remain
closed, precluding independent replication. Public survey-based datasets
(NLSY79/97~\cite{moore2000national}, CPS~\cite{cps2023}) offer nationally
representative samples but are limited in scale or temporal resolution.
Among recent public datasets, Decorte et al.~\cite{decorte2023career} released
$2{,}164$ career histories from anonymized Kaggle resumes mapped to
ESCO~\cite{le2014esco}, establishing the first public benchmark for
ESCO-standardized career-path prediction. Senger et al.~\cite{karrierewege}
constructed Karrierewege from $568{,}888$ career paths obtained from the German
Federal Employment Agency; occupations were already encoded in the Berufenet
taxonomy and mapped to ESCO, with free-text titles \emph{synthesized} by an LLM
rather than extracted from raw text. OpenResume~\cite{openresume} provides a
small complementary collection of $301$ real resumes augmented with synthetic
data. No prior public dataset simultaneously combines large scale, full
end-to-end extraction from unstructured text, rich temporal metadata,
education-level annotations, and ESCO standardization.

JobHop~\cite{jobhop-v1} addressed this gap as the first large-scale public
dataset to pair ESCO occupation codes with both temporal information and
educational attainment, derived end-to-end from real unstructured resumes.
Constructed from $361{,}207$ pseudonymized VDAB resumes, it contains $1.67$M work
experiences with quarter-level temporal annotations and a binary tertiary-degree
flag. JobHop~v2 retains this real-resume foundation while substantially improving
extraction quality, annotation breadth, and evaluation rigor.

\paragraph{LLM-based information extraction.}
In contrast to earlier rule-based and neural sequence-labeling extraction
pipelines~\cite{ayishathahira2018}, recent LLM-based approaches
have substantially narrowed the gap for unsupervised occupation extraction.
LLM4Jobs~\cite{llm4jobs} and SkillGPT~\cite{skillgpt} combine prompting with
vector-similarity search to extract and standardize occupations and skills
without task-specific supervision, while GoLLIE~\cite{sainz2024gollie} shows that
guideline-following instruction tuning substantially improves zero-shot
structured extraction under explicit annotation schemas. Within
career-trajectory pipelines, Decorte et al.~\cite{decorte2023career} used
GPT-3.5 to reformat experience sections into a uniform structure, and Senger et
al.~\cite{karrierewege} employed LLaMA~3.1 to synthesize free-text descriptions
from existing structured data. The JobHop setting is more demanding on several
axes: the input corpus consists of pseudonymized, multilingual (Dutch, French,
English) documents containing \texttt{<MASK>} tokens and heterogeneous
formatting, and because privacy constraints preclude cloud inference, the
extraction pipeline must run entirely on local infrastructure.

\section{The JobHop~v2 Dataset}\label{sec:dataset}

This section describes the JobHop~v2 dataset, from its underlying resume corpus
and ethical safeguards (Section~\ref{sec:data_source}) to the LLM-based
extraction pipeline that converts unstructured text into ESCO-coded career
trajectories (Section~\ref{sec:extraction}), the cleaning
(Section~\ref{sec:cleaning}) and normalization (Section~\ref{sec:normalization})
workflows applied to the raw output, and a summary of the dataset's key
statistics (Section~\ref{sec:dataset_summary}).

\subsection{Data Source and Ethical Considerations}\label{sec:data_source}

JobHop~v2 is derived from a corpus of ${\sim}440{,}000$ pseudonymized resumes
provided by VDAB, the Flemish Public Employment Service, under a formal research
collaboration agreement. Prior to sharing, VDAB converted all resumes to plain
text and applied systematic pseudonymization: personally identifiable
information and low-frequency terms were replaced with \texttt{<MASK>} tokens to
reduce indirect re-identification risk. The collection spans multiple languages
(predominantly Dutch, with smaller portions in French and English), reflecting
the multilingual labour market of the Flemish region of Belgium.

Data use is governed by the research agreement and complies with applicable
ethical and legal requirements. A binding constraint requires all processing to
occur on internal infrastructure: resume contents are never transmitted to
external services or third-party APIs. This constraint shaped the extraction
architecture (Section~\ref{sec:extraction}), which relies exclusively on locally
hosted model inference. The released dataset incorporates additional privacy
protections: location information is removed and all dates are coarsened to
quarter-level granularity, ensuring no personally identifiable information
remains in any public release. Documents whose formatting produced empty or
near-empty text after pseudonymization were excluded prior to extraction,
yielding approximately $400{,}000$ input resumes that entered the LLM extraction
stage.

\noindent\textit{Fairness note.}
Models trained on historical trajectories may perpetuate structural inequities in
labour-market access, occupational segregation, and systemic hiring
biases~\cite{de-artega2019bias,salinas2023unequal}. ESCO standardization provides
a neutral occupational vocabulary but does not eliminate the biases encoded in
the underlying trajectories, so any deployment should be accompanied by rigorous
fairness evaluation across demographic groups.

\subsection{LLM-Based Information Extraction}\label{sec:extraction}

The extraction pipeline uses a zero-shot, rule-rich instruction prompt with
strict structured JSON output. The prompt has two components. The \textbf{system
message} is a single sentence establishing the model's role (\textit{``You are an
expert HR assistant. Extract only what is explicitly present and format as valid
JSON.''}). The \textbf{developer instruction} contains the full rule set and
output schema, reproduced in Figure~\ref{fig:extraction_prompt}. It addresses
challenges characteristic of the corpus: pseudonymization artifacts
(\texttt{<MASK>} tokens), multilingual content, inconsistent date formats, and
heterogeneous multi-column formatting.

\begin{figure*}[align=\centering, pos={t!}]
\centering
\begin{minipage}[t]{0.54\textwidth}
  \vspace{0pt}
  \begin{tcolorbox}[
    enhanced, colback=blue!2, colframe=blue!50,
    colbacktitle=blue!12, coltitle=black,
    title=Extraction Rules (Sections 1--3), fonttitle=\small\bfseries,
    boxrule=0.5pt, arc=3pt,
    left=6pt, right=5pt, top=5pt, bottom=5pt,
    fontupper=\normalfont,
    equal height group=promptfig
  ]
  {\footnotesize\renewcommand{\ttdefault}{zi4}\ttfamily\raggedright
You are a strict Resume Parsing Engine. Your goal is to extract structured
data from the resume provided below into a single valid JSON object.

\medskip
\noindent\textbf{1.\ GENERAL CONSTRAINTS}\\
- \textbf{Input:} Resume text captured at timestamp: \{timestamp\}.\\
- \textbf{Output:} A single JSON object. No markdown, no explanations, no <MASK> tokens.\\
- \textbf{Language:} Do NOT translate descriptions or titles. Keep original language.\\
- \textbf{Missing Data:} Use ``-'' for missing/masked values. Drop parts containing only <MASK>.\\
- \textbf{Current Roles:} Replace markers like ``Present'', ``Heden'', ``Current'' with the\\
\quad provided timestamp formatted as ``Month Year''.

\medskip
\noindent\textbf{2.\ LOGIC \& PARSING RULES}\\
- \textbf{MERGE:} Consecutive entries for the same role/company with overlapping or adjacent\\
\quad dates must be combined into one entry.\\
- \textbf{SPLIT:} Single entries listing disjoint dates (e.g., ``2008, 2009 and 2015'') must be\\
\quad split into separate entries for each period.\\
- \textbf{GROUPS:} Unpack grouped sections (e.g., ``Interims'', ``Stages'') into individual experiences.\\
- \textbf{Dates:}\\
\quad Format: ``Month Year'' or ``Year''.\\
\quad Do not invent months. If only a year is listed, output only the year.\\
\quad Handle ranges: ``May July 2013'' $\to$ Start: ``May 2013'', End: ``July 2013''.\\
\quad Handle masks: ``<MASK>-2009'' $\to$ Start: ``-'', End: ``2009''.\\
\quad ``<MASK>/<MASK>/2015--<MASK>/<MASK>/2018'' $\to$ Start: ``2015'', End: ``2018''.\\
\quad \textbf{CLEANUP:} Ensure no hyphens or slashes remain from the mask.\\
\quad Always extract visible years.\\
- \textbf{Mappings:}\\
\quad Internships: ``stage'', ``trainee'', ``leerling'' $\to$ \texttt{contract\_type}: ``internship''.\\
\quad Degrees: ``Hogeschool'' $\to$ ``Bachelor''; ``Lic.'' $\to$ ``Master''; ``Doctoraat'' $\to$ ``PhD''.\\
\quad Languages: ``moedertaal/native'' $\to$ ``native''; ``vloeiend/fluent'' $\to$ ``fluent'';\\
\quad\quad ``goed'' $\to$ ``intermediate''; ``basis/notions'' $\to$ ``basic''.

\medskip
\noindent\textbf{3.\ EXTRACTION SCOPE}\\
- \textbf{Skills:} Extract ONLY technical skills/tools. Ignore soft skills.\\
- \textbf{Experience:}\\
\quad CRITICAL: Do NOT extract items from ``Education'' or ``Profile'' sections as work experience.\\
\quad Be careful with multi-column layouts.\\
\quad For the \texttt{title} field, extract the job title exactly as written.\\
\quad For the \texttt{standard\_title} field, infer a standardized title (preferably ESCO).\\
- \textbf{Certificates:} professional training/courses only.

  }
  \end{tcolorbox}
\end{minipage}
\hfill
\begin{minipage}[t]{0.45\textwidth}
  \vspace{0pt}
  \begin{tcolorbox}[
    enhanced, colback=blue!2, colframe=blue!50,
    colbacktitle=blue!12, coltitle=black,
    title=Output Schema (Section 4), fonttitle=\small\bfseries,
    boxrule=0.5pt, arc=3pt,
    left=6pt, right=5pt, top=5pt, bottom=5pt,
    equal height group=promptfig
  ]
  {\footnotesize\renewcommand{\ttdefault}{zi4}\ttfamily\begin{alltt}
\{
  \jkey{work\_experiences}: [
    \{
      \jkey{title}: "Original Job Title",
      \jkey{standard\_title}: "Standardized Job Title",
      \jkey{company\_name}: "Organization Name (or '-')",
      \jkey{location}: "City only (no country)",
      \jkey{description}: "Original text description",
      \jkey{start\_date}: "Month Year | Year | -",
      \jkey{end\_date}: "Month Year | Year | -",
      \jkey{work\_schedule\_type}:
        "full-time | part-time | -",
      \jkey{contract\_duration}:
        "temporary | permanent | seasonal | -",
      \jkey{contract\_type}:
        "employment | freelance | internship |
         volunteer | student job | -",
      \jkey{skills\_used}: ["Skill 1", "Skill 2"]
    \}
  ],
  \jkey{educations}: [
    \{
      \jkey{degree\_level}:
        "Primary education | Secondary school |
         Bachelor | Master | PhD",
      \jkey{degree\_title}: "Title (e.g. B.Sc. IT)",
      \jkey{institute\_name}: "School/University Name",
      \jkey{start\_date}: "Month Year | Year",
      \jkey{end\_date}: "Month Year | Year"
    \}
  ],
  \jkey{languages}: [
    \{
      \jkey{language}: "Name",
      \jkey{proficiency\_level}:
        "native | fluent | intermediate | basic"
    \}
  ],
  \jkey{other\_certificates}: [
    \{
      \jkey{certificate\_name}: "Name",
      \jkey{issuing\_organization}: "Issuer Name",
      \jkey{issue\_date}: "Month Year | Year"
    \}
  ]
\}
  \end{alltt}}
  \end{tcolorbox}
\end{minipage}
\caption{%
  Full developer instruction used for resume extraction.
  \textbf{Left:}~Extraction rules covering general constraints (\S1), date-handling logic
  and type-mapping conventions (\S2), and extraction scope (\S3).
  \textbf{Right:}~Target JSON output schema; field names shown in
  \textcolor{blue!60!black}{blue}.
  Curly braces denote JSON structure; \texttt{\{timestamp\}} is replaced at runtime
  with the capture date.
}
\label{fig:extraction_prompt}
\end{figure*}

\paragraph{Output schema.}
The schema defines four entity groups. \textbf{Work experiences} carry a job
title, an ESCO-aligned standardized title (a secondary signal for downstream
code assignment), company name, location, description, start and end dates, work
schedule (full- or part-time), contract duration (temporary, permanent, or
seasonal), contract type (employment, freelance, internship, volunteer, or
student job), and explicitly mentioned technical skills. \textbf{Education
entries} carry a degree level (PhD, Master, Bachelor, Secondary, Primary),
degree title, institution name, and start and end dates. \textbf{Languages}
carry a name and proficiency level (native, fluent, intermediate, or basic).
\textbf{Certificates} carry a name, issuing organization, and issue date.
This schema is substantially richer than the original JobHop scope, which was
limited to work experiences and a basic educational qualification; the language,
certificate, contract, work-schedule, technical-skill, and normalized
education-level fields are all new in v2. Technical skills are restricted to
tools and technologies explicitly present in the resume text; soft skills are
excluded by instruction to limit hallucination.

\paragraph{Inference configuration and throughput.}
Extraction runs on vLLM~\cite{kwon2023efficient} with
\texttt{openai/gpt-oss-120b}~\cite{openai2025gptoss}, a 120B-parameter reasoning
model, at \textsc{high} reasoning effort. The complete inference configuration is
listed in Table~\ref{tab:inference_params}. All computation was performed on a
single workstation with two NVIDIA H200~NVL GPUs ($140$\,GB HBM3e each), a
dual-socket AMD EPYC~9535 CPU, and $1.5$\,TiB of system RAM, satisfying the
on-premises processing constraint. A persistent retry loop reprocesses samples
failing JSON validation using an extended generation budget and an increased
repetition penalty: of ${\sim}400{,}000$ input resumes, ${\sim}50{,}500$
($12.6\%$) failed the initial pass and were all recovered through retry, yielding
a final corpus with a $100\%$ JSON parse rate. Figure~\ref{fig:extraction_example}
illustrates the end-to-end behavior of the pipeline on a synthetic resume
representative of the VDAB corpus.

\begin{table}[t]
\centering
\caption{Inference configuration for the production extraction pipeline.}
\label{tab:inference_params}
\small
\begin{tabular}{ll}
\toprule
\textbf{Parameter} & \textbf{Value} \\
\midrule
Model               & \texttt{openai/gpt-oss-120b} \\
Precision           & bfloat16 \\
Tensor parallelism  & 2 (one shard per GPU) \\
Prefix caching      & Enabled \\
Decoding            & Greedy ($T{=}0$) \\
Repetition penalty  & 1.1 (initial), 1.15 (retry) \\
Max generation tokens & 24{,}000 (initial), 32{,}000 (retry) \\
Reasoning effort    & \textsc{high} (both passes) \\
Batch size          & 200 resumes per batch \\
GPU memory utilization & 0.90 \\
\bottomrule
\end{tabular}
\end{table}

\begin{figure*}[align=\centering, pos={t!}]
\centering
\begin{minipage}[t]{0.44\textwidth}
  \vspace{0pt}
  \begin{tcolorbox}[
    enhanced, colback=gray!4, colframe=gray!50,
    colbacktitle=gray!20, coltitle=black,
    title={(a) Pseudonymized resume (input)}, fonttitle=\small\bfseries,
    boxrule=0.5pt, arc=3pt,
    left=6pt, right=5pt, top=5pt, bottom=5pt,
    lower separated=false,
    colbacklower=gray!8,
    fontupper=\normalfont,
    equal height group=examplefig
  ]
  {\footnotesize\renewcommand{\ttdefault}{zi4}\ttfamily\begin{alltt}
\textbf{Curriculum Vitae}

\hlmask{} \hlmask{}
\hlmask{}, 9000 Ghent
\hlmask{}@\hlmask{}.com   +32 \hlmask{}
Belgian, DOB \hlmask{}/1988

\textbf{WORK EXPERIENCE}

Sept. 2020 -- \textbf{Present}:
  Data Engineer
  \hlmask{} Solutions Inc., Brussels
  Responsible for ETL pipelines ...
  - Tools: Python, Spark, dbt

Jan. 2016 -- Aug. 2020:
  Junior Developer
  InnoSoft Ltd., Ghent
  - Back-end Java/Spring Boot; REST APIs
  - Tools: Java, Spring Boot, PostgreSQL

Mar. 2015 -- Dec. 2015:
  \textbf{Internship} Software Developer
  TechStart SA, Antwerp

\textbf{EDUCATION}
2013--2015  Master in Computer Science
            Ghent University
2010--2013  Bachelor in Applied Informatics
            University College Ghent

\textbf{LANGUAGES}
Dutch: \textbf{Native}
English: \textbf{Fluent} | French: \textbf{Good}

\textbf{CERTIFICATIONS}
Nov. 2021: AWS Certified Solutions Architect
           Amazon Web Services
\end{alltt}}
  \tcblower
  \footnotesize\textit{\textbf{Metadata}: Captured at February 2022}
  \end{tcolorbox}
\end{minipage}
\hfill
\begin{minipage}[t]{0.53\textwidth}
  \vspace{0pt}
  \begin{tcolorbox}[
    enhanced, colback=blue!2, colframe=blue!30,
    colbacktitle=blue!15, coltitle=black,
    title={(b) Structured JSON output (extracted)}, fonttitle=\small\bfseries,
    boxrule=0.5pt, arc=3pt,
    left=5pt, right=4pt, top=5pt, bottom=5pt,
    equal height group=examplefig
  ]
  {\footnotesize\renewcommand{\ttdefault}{zi4}\ttfamily\begin{alltt}
\{\jkey{work\_experiences}: [\{
   \jkey{title}: \jval{Data Engineer},
   \jkey{standard\_title}: \jval{Data engineer},
   \jkey{company\_name}: \jval{-},\jnote{masked \ensuremath{\to} \texttt{"-"}}
   \jkey{location}: \jval{Brussels},
   \jkey{start\_date}: \jval{Sept. 2020},
   \jkey{end\_date}: \jval{Feb. 2022},\jnote{Present \ensuremath{\to} timestamp}
   \jkey{description}: \jval{Responsible for...},
   \jkey{contract\_type}: \jval{employment},
   \jkey{skills\_used}: [\jval{Python}, \jval{Spark}, \jval{dbt}]
 \}, \{
   \jkey{title}: \jval{Junior Developer},
   \jkey{standard\_title}: \jval{Software developer},\jnote{ESCO-aligned}
   \jkey{company\_name}: \jval{InnoSoft Ltd.},
   \jkey{start\_date}: \jval{Jan. 2016},
   \jkey{end\_date}: \jval{Aug. 2020},
   \jkey{contract\_type}: \jval{employment},
   \jkey{skills\_used}: [\jval{Java}, \jval{Spring Boot}, ...]
 \}, \{
   \jkey{title}: \jval{Internship Software Developer},
   \jkey{standard\_title}: \jval{Software developer},
   \jkey{contract\_type}: \jval{internship},\jnote{inferred from title}
   \jkey{start\_date}: \jval{Mar. 2015},
   \jkey{end\_date}: \jval{Dec. 2015},
   \jkey{skills\_used}: [\jval{Django}]
 \}],
 \jkey{educations}: [\{
   \jkey{degree\_level}: \jval{Master},
   \jkey{degree\_title}: \jval{Master in Computer Science},
   \jkey{institute\_name}: \jval{Ghent University},
   \jkey{start\_date}: \jval{2013}, \jkey{end\_date}: \jval{2015}
 \}, \{
   \jkey{degree\_level}: \jval{Bachelor},\jnote{level from title}
   \jkey{degree\_title}: \jval{Bachelor in Applied Informatics},
   \jkey{institute\_name}: \jval{University College Ghent},
   \jkey{start\_date}: \jval{2010}, \jkey{end\_date}: \jval{2013}
 \}],
 \jkey{languages}: [
   \{\jkey{language}: \jval{Dutch},
    \jkey{proficiency\_level}: \jval{native}\},
   \{\jkey{language}: \jval{English},
    \jkey{proficiency\_level}: \jval{fluent}\},
   \{\jkey{language}: \jval{French},
    \jkey{proficiency\_level}: \jval{intermediate}\}\jnote{Good \ensuremath{\to} intermediate}
 ],
 \jkey{other\_certificates}: [\{
   \jkey{certificate\_name}: \jval{AWS Cert. Solutions Architect},
   \jkey{issuing\_organization}: \jval{Amazon Web Services},
   \jkey{issue\_date}: \jval{Nov. 2021}
 \}]\}
\end{alltt}}
  \end{tcolorbox}
\end{minipage}
\caption{%
  Extraction illustration using a synthetic resume representative of the VDAB corpus.
  For readability the example is shown in English; in practice resumes may be in any
  language, predominantly Dutch and French in our corpus, and the model preserves the
  original language rather than translating it.
  \textbf{(a)}~Pseudonymized input: \texttt{<MASK>}~tokens replace
  removed personally identifiable information.
  Bold terms mark surface cues that the prompt normalizes to controlled schema labels.
  \textbf{(b)}~Structured JSON produced by the extraction prompt, demonstrating five pipeline
  behaviors:
  (\emph{i})~masked fields resolved to~\texttt{"-"};
  (\emph{ii})~present-tense marker \emph{Present} resolved to the capture timestamp;
  (\emph{iii})~free-text descriptions extracted and preserved (shown truncated);
  (\emph{iv})~job titles mapped to ESCO-aligned \texttt{standard\_title} values;
  (\emph{v})~surface terms normalized to controlled schema labels
  (\emph{Internship}~$\to$~\texttt{internship}, \emph{Good}~$\to$~\texttt{intermediate}).
}
\label{fig:extraction_example}
\end{figure*}

\subsection{Data Cleaning Pipeline}\label{sec:cleaning}

The raw JSON outputs pass through a five-stage cleaning pipeline, applied
sequentially, each operating on the output of the previous one.

\begin{enumerate}
  \item \textbf{Parsing-artifact correction.} Dates containing extraction
    artifacts are identified via regular expressions targeting residual
    \texttt{<MASK>} tokens within date fields, future-year placeholders (years
    $> 2025$), and malformed strings with dangling hyphens or slashes. Affected
    dates are corrected to the extractable year component or set to missing.
  \item \textbf{Invalid-range correction.} Entries where the parsed start date is
    later than the end date are flagged. Where the discrepancy is small ($\leq 1$
    quarter) the dates are swapped; entries with implausible durations ($> 40$
    years) are removed.
  \item \textbf{Exact deduplication.} Entries identical across all core fields
    (title, company, start date, end date) within the same resume are collapsed
    to a single entry, addressing artifacts where multi-column layouts cause the
    same position to be extracted twice.
  \item \textbf{Quality filtering.} Resumes are removed if they contain more than
    $20$ work-experience entries (likely parsing errors on non-resume documents)
    or zero meaningful entries across all entity groups after earlier stages.
  \item \textbf{Consecutive-job merging.} Adjacent entries within the same resume
    are merged if they share the same normalized job title, the same normalized
    company name, and temporally contiguous or overlapping date ranges (gap
    $\leq 1$ quarter); descriptions are concatenated. This stage consolidates
    approximately $19{,}432$ entries split during extraction due to multi-line or
    multi-column resume formatting.
\end{enumerate}

\subsection{Normalization Pipeline}\label{sec:normalization}

After cleaning, the extracted fields are normalized into controlled
representations suited to cross-resume analysis and modeling. Normalization
comprises three independent components: assigning ESCO occupation codes to job
entries, converting dates to quarter-year granularity, and mapping education
entries to a common degree-level taxonomy.

\paragraph{Occupation-code assignment.}
Individual job titles are highly variable, so cross-resume analysis requires
mapping them to a controlled vocabulary~\cite{llm4jobs}. We adopt the ESCO
taxonomy (v1.1.2)~\cite{le2014esco}, organized hierarchically with up to
$3{,}007$ leaf-node occupations. Occupation-code assignment uses a commercial
classifier applied via a two-step hybrid policy that exploits the richer
JobHop~v2 extraction schema. \emph{Primary assignment} passes the concatenation
of the extracted \texttt{title} and \texttt{description} fields to the
classifier, accepting the top-1 prediction if its confidence exceeds
$\tau_1 = 0.45$. \emph{Rescue assignment} submits the model-generated
\texttt{standard\_title} for entries below $\tau_1$, accepting the top-1
prediction only if confidence exceeds a stricter threshold $\tau_2 = 0.85$.
Entries unassigned after both steps are labeled \texttt{unknown}. The asymmetric
thresholds are deliberate: primary assignment maximizes coverage at moderate
confidence, whereas rescue assignment acts as a high-precision fallback activated
only when the standardized title provides a strong signal.

\paragraph{Temporal normalization.}
All dates are converted to quarter-year format (\texttt{Q\# YYYY}): a
multilingual mapping first converts Dutch, French, and English month names to a
canonical numeric representation; regular expressions then extract year and
optional month from heterogeneous formats (``MM-YYYY'', ``YYYY-MM'',
``Month YYYY'', abbreviated forms such as ``jan.\ 2015''); months are mapped to
quarters (1--3~$\to$~Q1, \dots, 10--12~$\to$~Q4, month-only dates default to Q1);
and years outside $[1950, 2025]$ are rejected.

\paragraph{Education-level normalization.}
Education entries are mapped to a five-level taxonomy via a multilingual
term mapping: \emph{PhD} (Doctoraat, PhD, Doctorate); \emph{Master}
(Master, Licentiate, Lic., Ma.); \emph{Bachelor} (Bachelor, Hogeschool,
Professional Bachelor, Ba.); \emph{Secondary} (Secondary, Middelbaar,
Secondaire, ASO, TSO, BSO); and \emph{None} (Primary, Lager onderwijs,
Primaire, and other degrees). When the \texttt{degree\_level} field is
missing or uninformative, keyword matching is applied to the
\texttt{degree\_title} field, recovering degree-level information for
approximately $3{,}194$ entries. The highest attained degree level per resume is
computed by taking the maximum across all education entries.

\subsection{Dataset Summary}\label{sec:dataset_summary}

After extraction, normalization, and cleaning, JobHop~v2 comprises $355{,}315$
unique resumes, $1{,}993{,}291$ work-experience entries, and $923{,}981$
education entries.\footnote{\ifanonymous
\else The dataset is publicly available at
\url{https://huggingface.co/datasets/aida-ugent/JobHop}, and the code at
\url{https://github.com/aida-ugent/Step}.\fi}
ESCO assignment is dominated by the primary title-plus-description path
($93.0\%$); the rescue mechanism adds $1{,}484$ labels ($0.1\%$); the remaining
$6.9\%$ are labeled \texttt{unknown}, reflecting deliberate conservatism to
prioritize label quality over coverage.

Compared with prior public benchmarks, JobHop~v2 remains distinctive in being
constructed through full end-to-end extraction from real, unstructured
resumes, rather than from pre-standardized occupational codes or
LLM-synthesized text. Decorte et al.~\cite{decorte2023career} extracted $2{,}164$
histories from Kaggle resumes at considerably smaller scale and without temporal
metadata or education annotations;
OpenResume~\cite{openresume} releases only $301$ real English trajectories; and
Karrierewege~\cite{karrierewege} provides $568{,}888$ paths built from Berufenet
occupational codes rather than raw resume text, with free-text descriptions
synthesized rather than extracted.

\begin{figure*}[align=\centering, pos={t}]
  \centering
  \begin{subfigure}[b]{0.58\textwidth}
    \centering
    \includegraphics[width=\linewidth]{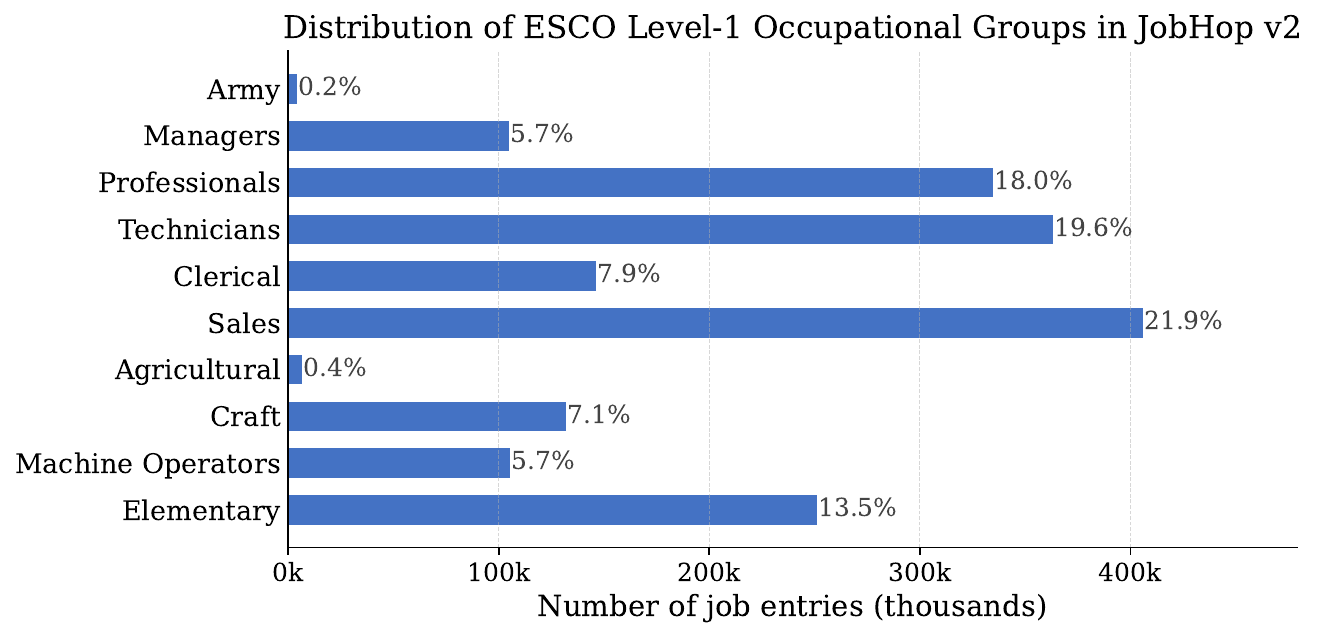}
    \caption{ESCO level-1 distribution.}
    \label{fig:esco_distribution}
  \end{subfigure}
  \hfill
  \begin{subfigure}[b]{0.38\textwidth}
    \centering
    \includegraphics[width=\linewidth]{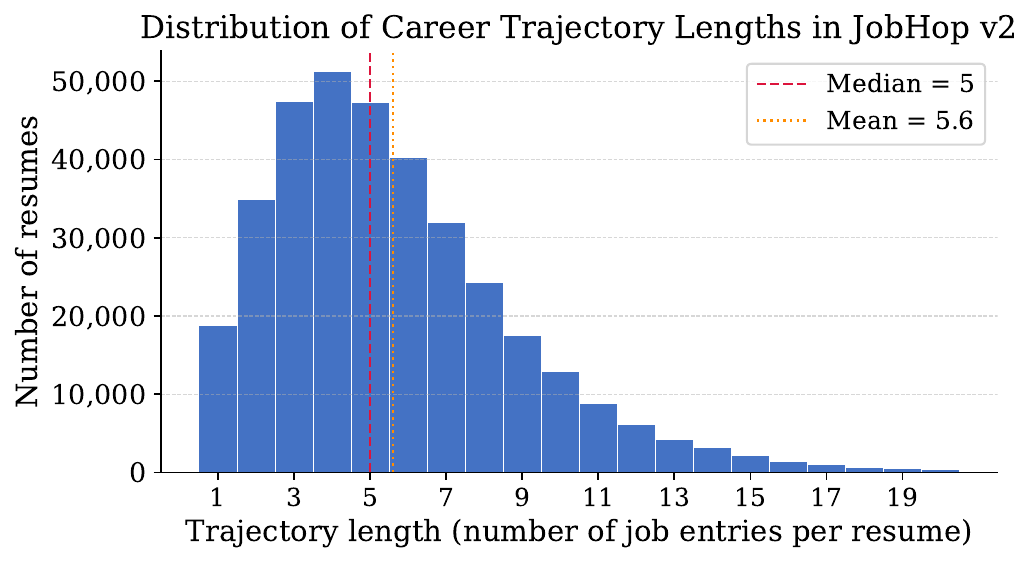}
    \caption{Trajectory lengths.}
    \label{fig:trajectory_lengths}
  \end{subfigure}
  \caption{Occupational diversity and career-history depths in JobHop~v2.
    Trajectory lengths range from 1 to over 20 jobs, with a median of~5.}
  \label{fig:dataset_distributions}
\end{figure*}

Figure~\ref{fig:dataset_distributions} characterizes the dataset along two
complementary dimensions. The occupational distribution
(Figure~\ref{fig:esco_distribution}) shows that Service~\&~Sales Workers (ISCO~5),
Professionals (ISCO~2), and Technicians~\&~Associate Professionals (ISCO~3)
account for the largest share of entries, reflecting the white-collar composition
of the underlying resume corpus; elementary occupations (ISCO~9) are next most
frequent, while primary-sector occupations (ISCO~6) are rare. The
trajectory-length distribution (Figure~\ref{fig:trajectory_lengths}) is
right-skewed: most individuals have between two and six recorded positions, but a
non-trivial fraction contain ten or more entries, providing rich sequential
context for downstream prediction models. Together, the two panels confirm that
JobHop~v2 captures a broad occupational spectrum and a realistic range of
career-history depths, supporting both occupational embedding evaluation and
next-occupation prediction.

\section{Extraction Evaluation}\label{sec:extraction_eval}

Reliable evaluation of LLM-based extraction requires both a scoring protocol that
faithfully reflects quality for downstream use and reference annotations that
minimize labeling noise as a confound. Section~\ref{sec:eval_protocol} describes
the revised evaluation protocol; Section~\ref{sec:eval_baselines} the three
labeled reference sets; and Section~\ref{sec:eval_results} the model comparison.

\subsection{Evaluation Protocol}\label{sec:eval_protocol}

We evaluate extraction quality on $200$ benchmark resumes from the original
JobHop annotation study~\cite{jobhop-v1}, using a revised protocol that replaces
binary date scoring and unweighted text matching with graduated partial-credit
scoring, length-ratio-gated string comparison, and field-importance weighting.
The protocol normalizes each field, scores it with a string-similarity or
date-specific function, weights fields by importance, and composes these into a
single sample-level score.

\paragraph{String similarity.}
Operating on normalized text (lowercased, whitespace-collapsed, abbreviations
unified), the similarity function returns $100$ for an exact match or two
empty fields, $10$ when one field is meaningful and the other missing, a
length-ratio-gated score for substring containment ($90$/$75$/$50$ for length
ratio $r > 0.7$, $0.5 < r \leq 0.7$, $r \leq 0.5$), and a graduated
Levenshtein-based score otherwise (full credit at ratio $s \geq 90$, decreasing
penalties below).

\paragraph{Date scoring.}
Dates are parsed flexibly, supporting English, Dutch, and French month names,
with present/ongoing markers mapped to a reference date. Scores range from $85$
(year and month match) and $60$ (year-only match) down to $15$ (no match), with
intermediate values for partial parses and missing fields.

\paragraph{Field-importance weights.}
Fields are weighted by downstream importance: Title and Company/Institute name
$1.5\times$; Start and End dates $1.3\times$; Description $1.2\times$;
Place/Location and Type $1.0\times$; Degree level $0.9\times$; other fields
$1.0\times$ by default.

\paragraph{Sample-level composition.}
The overall score per sample is
\begin{equation*}
  S = P_{\text{count}} \cdot
      \bigl(0.6 \cdot \bar{M} + 0.2 \cdot R_{\text{entry}}
             + 0.2 \cdot P_{\text{entry}}\bigr),
\end{equation*}
where $\bar{M}$ is the mean matched-entry weighted similarity (computed via the
Hungarian algorithm), $R_{\text{entry}}$ and $P_{\text{entry}}$ are entry-level
recall and precision, and $P_{\text{count}}$ is an entry-count penalty,
\begin{equation*}
  P_{\text{count}} = \max\bigl(0.5,\; 1.0 - 0.3 \cdot
    |n_{\text{gt}} - n_{\text{pred}}| /
    \max(n_{\text{gt}}, n_{\text{pred}})\bigr).
\end{equation*}

\subsection{Labeled Reference Datasets}\label{sec:eval_baselines}

The $200$ benchmark resumes were annotated by \ifanonymous
\else the original JobHop authors and external annotators\fi{} over
approximately $60$ cumulative hours. During
development of the revised protocol we observed that these annotations contain
occasional inconsistencies (divergent date-field granularity, differing
conventions for ambiguous section boundaries, subjective judgment calls) that can
systematically bias accuracy estimates. To disentangle genuine extraction errors
from annotation noise, we evaluate against three annotation sets:
\begin{enumerate}
  \item \textbf{Hand annotations}: the original multi-annotator labels,
    reflecting genuine human judgment but exhibiting inter-annotator variability.
  \item \textbf{Edited labels}: a systematic revision by an AI agent (Claude
    Sonnet) instructed to resolve inconsistencies and enforce uniform conventions
    while preserving correct annotations; lower noise, same structure.
  \item \textbf{Agent labels}: an independent annotation set generated by the
    same AI agent \emph{without access} to the hand or edited annotations,
    providing an internally consistent reference unanchored to the conventions or
    errors of the original process.
\end{enumerate}
These form a progression in annotation consistency, from multi-annotator hand
labels (highest variability), through edited labels (reduced noise), to agent
labels (most internally consistent). Evaluating against all three assesses
whether comparative conclusions are robust to the choice of reference.

\subsection{Results}\label{sec:eval_results}

Five extraction systems are compared under the revised protocol: GPT-OSS-120B
with \textsc{high} and \textsc{medium} reasoning, GPT-OSS-20B,
Llama-3.3-70B~\cite{llama3}, and Gemma-2~\cite{gemma2}. GPT-OSS-120B is evaluated
across $10$ independent extraction runs; $95\%$ confidence intervals are computed
via the $t$-distribution with $9$ degrees of freedom.

\begin{table*}[t]
\centering
\caption{Extraction accuracy (\%) across the three labeled reference sets.
\textbf{Sim.}: weighted text similarity ($60\%$ component weight).
\textbf{Rec.}/\textbf{Prec.}: entry-level recall/precision ($20\%$ each).
\textbf{Final}: full weighted score after the entry-count penalty.}
\label{tab:ie_model_comparison}
\resizebox{\textwidth}{!}{%
\begin{threeparttable}
\begin{tabular}{lcccccccccccc}
\toprule
& \multicolumn{4}{c}{\textbf{Hand Ann.}} & \multicolumn{4}{c}{\textbf{Edited}} & \multicolumn{4}{c}{\textbf{Agent}} \\
\cmidrule(lr){2-5} \cmidrule(lr){6-9} \cmidrule(lr){10-13}
\textbf{Model} & \textbf{Sim.} & \textbf{Rec.} & \textbf{Prec.} & \textbf{Final} & \textbf{Sim.} & \textbf{Rec.} & \textbf{Prec.} & \textbf{Final} & \textbf{Sim.} & \textbf{Rec.} & \textbf{Prec.} & \textbf{Final} \\
\midrule
Gemma-2~\cite{gemma2}          & 79.4 & 96.2 & 92.6 & 81.6 & 80.3 & 95.6 & 97.1 & 84.8 & 83.1 & 96.6 & 95.7 & 86.0 \\
Llama-3.3-70B~\cite{llama3}    & 79.1 & \textbf{97.2} & 93.9 & 82.5 & 79.0 & \textbf{96.5} & 97.7 & 84.8 & 83.4 & \textbf{97.8} & 95.7 & 86.6 \\
GPT-OSS-20B                    & 78.9 & 94.9 & 94.6 & 82.0 & 78.2 & 93.6 & 97.8 & 83.2 & 82.7 & 95.1 & 96.3 & 85.5 \\
GPT-OSS-120B (\textsc{med.})   & 80.3 & 94.7 & \textbf{95.7} & 83.3 & 79.3 & 93.5 & \textbf{98.9} & 84.4 & 84.2 & 94.9 & \textbf{97.0} & 86.7 \\
GPT-OSS-120B (\textsc{high})   & \textbf{81.9} & 96.4 & 93.9 & \textbf{83.9} & \textbf{81.7} & 96.0 & 97.6 & \textbf{86.2} & \textbf{86.4} & 97.0 & 95.5 & \textbf{88.0} \\
\midrule
Label agreement\tnote{a}       & 84.4 & 96.6 & 95.8 & 86.6 & 86.2 & 97.1 & 96.7 & 88.5 & 87.4 & 97.0 & 96.6 & 89.1 \\
\bottomrule
\end{tabular}
\begin{tablenotes}[flushleft]
\scriptsize
\item Similarity, Recall, and Precision are sample-level component means under the revised protocol; Final is the weighted score after the entry-count penalty.
\item GPT-OSS-120B values are means over $10$ independent runs ($\pm 0.2$--$0.5$~pp $95\%$ CI). Bold = best model result per column (label agreement excluded).
\item[a] Symmetric pairwise agreement between label sets, providing an interpretive ceiling.
\end{tablenotes}
\end{threeparttable}%
}
\end{table*}

Table~\ref{tab:ie_model_comparison} presents the results. The \emph{Label
agreement} row reports the symmetric pairwise agreement of each reference set with
the other two, and acts as the \emph{annotator-agreement ceiling} for its column:
because the human and agent labels themselves disagree at this level, no extractor
should be expected to exceed it, and the objective for each model is to come as
close to it as possible. Read this way, GPT-OSS-120B at \textsc{high} reasoning is
the strongest system: it comes closest to the ceiling on all three reference sets,
at $83.9$ vs.\ $86.6$ (Hand), $86.2$ vs.\ $88.5$ (Edited), and $88.0$ vs.\ $89.1$
(Agent), a gap of only $1.1$--$2.7$~percentage points, and does so by a smaller margin
than any other model, while also outperforming \textsc{medium} reasoning by
$0.6$--$1.8$~pp, which justifies the additional inference cost. Llama-3.3-70B
performs competitively, approaching GPT-OSS-120B (\textsc{medium}) on agent
labels; Gemma-2 achieves comparable accuracy on edited and agent labels despite a
lower JSON parse rate (${\approx}86\%$ vs.\ ${\geq}98.5\%$), indicating that its
errors concentrate in format compliance rather than extraction quality.

That the best extractor sits within $1.1$--$2.7$~pp of the agreement ceiling
suggests that much of the residual error reflects genuine annotation ambiguity
rather than systematic model failure. Reported accuracy also increases from hand
annotations through edited to agent labels for every model; this is consistent
with the intended progression in annotation consistency, though we note that the
agent labels were themselves produced by an LLM and may share inductive biases
with the extractors, so scores against them should not be read as a fully
independent measure of quality.

\subsection{Head-to-Head Comparison against JobHop~v1}\label{sec:eval_pairwise}

The evaluation above measures agreement with fixed annotations on the $200$
benchmark resumes. To test whether the redesigned pipeline yields better
extractions than the original JobHop pipeline on the corpus at large, we
additionally run a blind pairwise comparison judged by an LLM.

\paragraph{Protocol.}
For each of $1{,}000$ resumes sampled from the shared corpus, we render the
JobHop~v1 and JobHop~v2 extractions into a common plain-text schema exposing only
the fields present in \emph{both} releases (work experiences, education, and
certificates), so that neither side is identifiable by formatting; fields unique
to v2 (standardized titles, skills, contract and work-schedule type) are omitted.
The two extractions are presented as ``Extraction~A'' and ``Extraction~B'' in a
randomized order, alongside the \emph{original} resume as reference. A judge model
(Kimi~K2.6) scores each extraction on \emph{accuracy} (fidelity of titles,
companies, dates, and descriptions to the resume) and \emph{completeness} (whether
all experiences, educations, and certificates are captured) on a $0$--$10$ scale,
with accuracy weighted more heavily and pure formatting differences disregarded;
it also selects an overall winner. The presentation order is remapped afterward
for analysis.

\begin{table}[t]
\centering
\caption{Blind pairwise comparison of JobHop~v2 vs.\ JobHop~v1 extractions,
judged by Kimi~K2.6 over $1{,}000$ resumes with the original resume as reference.
Win rates (\%) and mean judge scores ($0$--$10$); bold marks the preferred
release.}
\label{tab:pairwise}
\small
\begin{tabular}{lccc}
\toprule
& \textbf{JobHop~v2} & \textbf{JobHop~v1} & \textbf{Tie} \\
\midrule
\multicolumn{4}{l}{\emph{Win rate}} \\
\quad Overall      & \textbf{68.3} & 29.9 & 1.8 \\
\quad Work exp.    & \textbf{59.7} & 29.4 & 10.9 \\
\quad Education    & 33.4 & 31.2 & 35.4 \\
\midrule
\multicolumn{4}{l}{\emph{Mean score}} \\
\quad Accuracy     & \textbf{8.60} & 7.81 & -- \\
\quad Completeness & \textbf{8.24} & 7.38 & -- \\
\bottomrule
\end{tabular}
\end{table}

\paragraph{Results.}
Table~\ref{tab:pairwise} summarizes the outcome. JobHop~v2 is preferred overall
on $68.3\%$ of resumes versus $29.9\%$ for v1 ($1.8\%$ ties; $69.6\%$ vs.\
$30.4\%$ excluding ties), a highly significant margin. The advantage is
concentrated in work experiences (v2 preferred on $59.7\%$ vs.\ $29.4\%$), whereas
education is effectively at parity (v2 $33.4\%$ vs.\ v1 $31.2\%$, $35.4\%$ ties),
consistent with the schema and prompt changes chiefly targeting work-experience
fields. Mean judge scores favor v2 on both axes (accuracy $8.60$ vs.\ $7.81$;
completeness $8.24$ vs.\ $7.38$). The preference is invariant to presentation
order (v2 wins $66.0\%$ and $70.3\%$ of decisions depending on its slot), and the
judge's declared winner agrees with the sign of its score difference on $99.8\%$
of resumes, indicating internally consistent verdicts. The judge's free-text
rationales most often attribute v2's advantage to cleaner handling of redacted
(\texttt{<MASK>}) text, certificate capture, and multilingual date normalization.

\subsection{Limitations}\label{sec:limitations}

Several limitations qualify these results. First, occupational coverage is
deliberately conservative: $6.9\%$ of work experiences remain \texttt{unknown}
because the two-step assignment policy favors label quality over completeness.
Second, the pairwise comparison relies on a single judge model; although its
verdicts are order-invariant and internally consistent, they may still encode
model-specific preferences, and the observed effect size is only small-to-moderate
($d \approx 0.48$) despite the very small $p$-values afforded by the large sample.
Third, the pairwise judge sees only the schema fields shared by both releases, so
the additional v2 fields (skills, languages, contract and work-schedule type) do
not contribute to the comparison, which likely \emph{understates} v2's added
value. Finally, the reference-based evaluation uses $200$ resumes from the
original annotation study, and both evaluations are confined to the VDAB corpus
and its three languages, so generalization to other labour markets remains to be
established.

\section{Conclusion}\label{sec:conclusion}

We presented JobHop~v2, an improved large-scale career-trajectory dataset
constructed through end-to-end LLM-based extraction from a corpus of
${\sim}440{,}000$ pseudonymized multilingual resumes provided by VDAB. The
released dataset
comprises $355{,}315$ unique career trajectories annotated with ESCO occupational
codes, quarter-level temporal information, and normalized five-level education
attainment. It improves on the original JobHop release in extraction quality,
evaluation rigor, and annotation breadth, while preserving its public,
real-resume foundation.

The accompanying extraction pipeline, based on reasoning-controlled LLM inference
with a multi-step retry mechanism, achieves a $100\%$ JSON parse rate after retry
and, across three annotation baselines, comes closest of all compared models to
the inter-annotator agreement ceiling, trailing it by only $1.1$--$2.7$~pp. In a
blind pairwise evaluation with the original resume as reference, an LLM judge
preferred JobHop~v2 extractions over the original JobHop pipeline on $68.3\%$ of
resumes versus $29.9\%$. The revised evaluation protocol, with graduated
partial-credit scoring and three complementary annotation baselines, provides a
more faithful assessment of extraction quality than prior approaches.

JobHop~v2 is intended as a shared resource for the career-trajectory research
community. Immediate applications include next-occupation prediction,
occupational embedding evaluation, and labour market analysis. Future directions
include extending the extraction pipeline to additional labour markets and
languages, and supporting fairness evaluation across education levels and
occupational sectors.

\ifanonymous
\else
\begin{acknowledgments}
  This research was funded by the BOF of Ghent University (BOF20/IBF/117), the
  Flemish Government (AI Research Program), the FWO (G073924N), and the EU (ERC,
  VIGILIA, 101142229). Views and opinions expressed are however those of the
  author(s) only and do not necessarily reflect those of the EU or the ERC
  Executive Agency. Neither the EU nor the granting authority can be held
  responsible for them. For the purpose of Open Access the author has applied a
  CC BY public copyright license to any Author Accepted Manuscript version
  arising from this submission. Part of the experiments were conducted on
  pseudonimized HR data generously provided by VDAB.
\end{acknowledgments}
\fi

\section*{Declaration on Generative AI}
During the preparation of this work, the author(s) used GPT-OSS-120B, Claude
Sonnet, and Kimi~K2.6 in order to: build the dataset extraction, annotation, and
evaluation pipeline described in the paper (a core research contribution), and,
separately, for grammar and spelling checking of the manuscript. After using these tools, the
author(s) reviewed and edited the content as needed and take(s) full
responsibility for the publication's content.

\bibliography{bibliography}

\end{document}